\documentclass[conference]{IEEEtran}
\usepackage{cite}
\usepackage[]{algorithm2e}
\usepackage{graphicx}
\graphicspath{ {images/} }

\begin{document}

%Here goes the title

\title{Occlusion-Based Cooperative Transport for Concave Objects with a Swarm of Miniature Mobile Robots}

%Authors List

\author
{\IEEEauthorblockN{Sanjuksha Nirgude, Animesh Nema, Aishwary Jagetia}
\IEEEauthorblockA{ Robotics Engineering Department\\
Worcester Polytechnic Institute\\
Worcester, MA, USA\\
snirgude@wpi.edu, anema@wpi.edu, adjagetia@wpi.edu}
}
\maketitle

%Main body starts

\begin{abstract}
An occlusion based strategy for collective transport of a concave object using a swarm of mobile robots has been proposed in this paper. We aim to overcome the challenges of transporting concave objects using decentralized approach. The interesting aspect of this task is that the agents have no prior knowledge about the geometry of the object and do not explicitly communicate with each other. The concept is to eliminate the concavity of the object by filling a number of robots in its cavity and then carry out an occlusion based transport strategy on the newly formed convex object or "pseudo object". We divide our work into two parts- concavity filling of various concave objects and occlusion based collective transport of convex objects.   
\end{abstract}

\begin {IEEEkeywords}
decentralized approach, collective transport, occlusion based transport, concave object, convex object.
\end{IEEEkeywords}

\section{Introduction and Background}
\label{intro}
Swarm Intelligence is the collective behavior of decentralized, self-organized systems which could be either natural or man-made. Swarm Intelligence exists in nature such as the behavior of ants, bees, birds which can be used as an inspiration for finding its applications in the field of robotics. One such application is the transportation of objects using a number of mobile robots. 
Such tasks may seem trivial at first, but can be incredibly complicated, depending on various aspects such as shape of the object, size of the object, visibility and perception. In addition to that, the process is decentralized i.e., every agent behaves independently instead of following a fixed leader. 
There are various techniques to transport an object using decentralized approach. General structure of a robot's behavior, performing collective transport consists of searching the object, positioning itself around the object and then transporting the object\cite{c4}.

Collective transport methods can be categorized into three categories: Pulling, Pushing and Caging. Pulling constitutes of complex mechanisms like grasping and lifting the objects, whereas caging requires robots to maintain their formation during dynamic movements. During pushing, robot's pushing positions and speed are the constraints to be addressed. Increasing the number of pushing robots increases the stability of the object as pushing force is distributed over multiple points. Also, due to hardware requirements for pulling and caging strategies, we have preferred pushing strategy for our analysis. 

In the paper\cite{c4} a simple odometry based co-ordination strategy in combination with omni-directional camera has been used. There is no communication between robots while performing collective transport task. The transportation strategy in this paper consists of four stages, namely prey discovery, team co-ordination, recruitment and transportation. In \cite{c2} a collective transport approach has been proposed by the authors using kilobots and r-ones. The agents have no prior knowledge about the object's shape,size, location of its neighbors or the object. They only know the location of the goal. The agents perceive the direction of the goal by using their light sensor(s) and apply forces on the object in the direction of perceived light. By doing so, they optimally transport objects of complex shapes to the desired location. The r-ones also execute flocking behavior in case some agents are occluded from the light source. In such a case, the robots observe the direction of their neighbors to modify their own directions. The authors proved experimentally the scalability, robustness and optimality of their approach by testing their agents under different circumstances. In \cite{c3} the authors have proposed a 5 step approach to successfully transport an object of any shape be it, concave or convex. The tasks assigned to the robots are in the following order. First the robots have to explore the environment and locate the object. Once the robots find the object, they will align themselves with the object and grasp it. The third step is object characterization. i.e., finding the centroid of the object,width,diameter, orientation etc. They do so by determining their own positions and centroid. The object’s diameter determines the minimum distance from an obstacle where it is safe to rotate the object. Once the object information has been extracted, the robots then perform a path planning function. During this phase, some of the robots stay attached to the object while others explore a suitable path. The robots then navigate the object through a chosen path. The paper talks about various algorithms for each of it's steps, hence giving an idea about collective transport of complex objects. 

Our paper uses an occlusion based collective transport strategy.The concept behind the occlusion based strategy is that each agent searches the object, moves towards it and then looks for the goal. All the robots push the object by moving in a direction perpendicular to the object’s surface at their points of contact. This way the motion of the object will be approximately towards the goal. The robots work in a decentralized manner and conduct co-operative transport without explicitly communicating with each other. Our project is divided into two parts in which we fill the concavity of the objects and also complete the collective transport of convex object separately. In case of collective transport experiment the location of the object is known by the robot along with its own position.In the collective transport experiment the task sequence starts with the robot orienting towards the object and moving towards it. When the robot reaches the object, it checks if the goal in our case the light source is visible from its position, if the goal is visible,it moves around the object and looks for the goal again. If the goal is not visible (or occluded), the robot starts pushing the object. Otherwise the robots moves around object, executing a left-hand-wall-following behavior. This is repeated by every agent collectively, till the object is transported to the goal. While this approach has been successfully tested on convex objects \cite{c1}, concave objects pose much harder challenges. The agents can easily lose the sense of direction towards the goal and might never transport the object.

Our motivation for this project comes from the paper \cite{c1}. In which one of the major limitation was its inability to transport concave objects. We propose a method to overcome this drawback of the occlusion based strategy. The object's concavity can be eliminated by filling a number of agents in it's concave contour, thus making it more 'convex' like. Another swarm of agents which will be referred to as the 'pushing agents' can then perform an occlusion based approach mentioned above by treating the object as any other convex polygon. Therefore, a concave object can be transported to the goal. 

\section{Proposed Work}
\label{section2}

\begin{figure}[ht]
  \centering
  \includegraphics[width = 8cm]{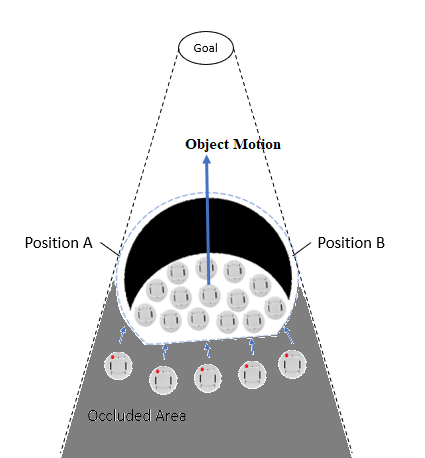}
   \caption{Occlusion-Based Collective Transport for Concave Object}
   \label{fig.1}
\end{figure}

\begin{figure}[!ht]
  \centering
  \includegraphics[width = 8cm]{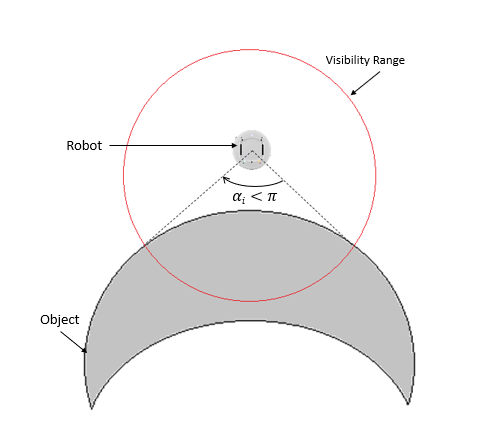}
   \caption{When robot approaches from convex side}
   \label{fig.2}
\end{figure}

\begin{figure}[ht]
  \centering
  \includegraphics[width = 8cm]{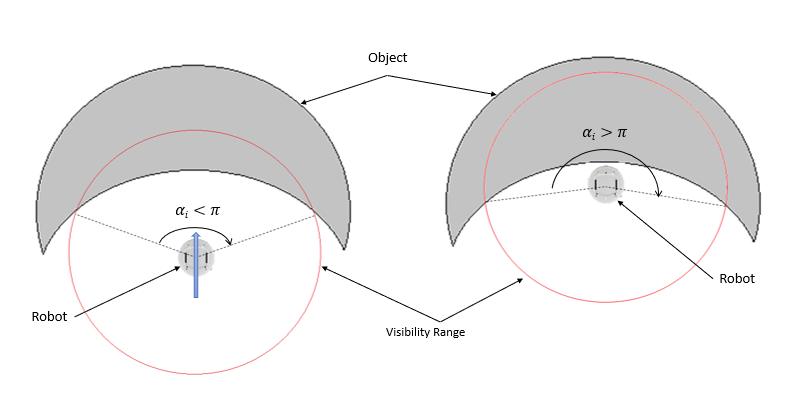}
   \caption{When robot approaches from concave side}
   \label{fig.3}
\end{figure}

As described above, the occlusion based collective transport strategy only works when the object is in convex shape. If the object is concave the strategy fails. Based on our strategy, initially the robot searches for an object while performing a random walk. Once the object is seen, robot approaches the object. When the robot reaches the object, it performs a left-hand-wall-following behavior around the object to checks if the object is concave or convex. If the range of the angles where the object is detected relative to each robot is greater than \(pi\), the object is identified to be a concave object, else it is a convex object. 
It can be seen in the figure \ref{fig.2} that the robot is on the convex side of the object. Hence, the angle is less than \(pi\) and the object will be identified as convex object. Whereas in figure \ref{fig.3} it can be seen that the robot senses the object at an angle more than \(pi\) ,therefore identifying it as an concave object.

If the object is detected to be concave, the robot will stay at the particular position and change its robot id to "\textbf{object robots}". Hence, filling the concavity of the object. When a next robot arrives behind this robot it will consider the first robot as an object and check if the concavity of the object still exists. This process will repeat until the concavity of the object is filled and robots collectively form a "\textbf{pseudo object}" which incorporates both the actual object and the object robots. Once the pseudo object changes its shape to convex, the remaining robots will change their robot id to "\textbf{pushing robots}", and the "object robots" will stay intact with the object. Hence, the task of converting the concave object to convex object will be accomplished.

At this point, the occlusion based collective transport strategy comes into effect. Now the robot checks if the goal can be seen from its position.  If it is not visible or if the robot is in occluded region the robot must start pushing the object. Otherwise the robots should move around object, executing a left-hand-wall-following behavior. 

As shown in figure \ref{fig.3} the robots after filling the concavity of the object now start applying the occlusion based collective transport strategy for moving the object. The robots with red LED's are now the "pushing robots" in the occluded area of the object. The position A and the position B are the positions of the extreme observer robots from where the goal is visible.

\section{Experiments}
\label{section3}
Our Experiments are broadly divided into two sub-experiments. We have implemented the concave filling experiment in Buzz programming language \cite{c5} and collective transport experiment in C++. We have simulated both the experiments in ARGoS Simulator \cite{c6}.  

\subsection{Experiment 1: Concave Filling}
Concave filling for different concave shapes is shown in the figure \ref{fig.4}. Our environment setup is shown in \ref{fig.7}, \ref{fig.14} and \ref{fig.15} for different shapes.

\subsubsection{Experimental Setup}
\begin{figure}[ht]
  \centering
  \includegraphics[width = 8cm]{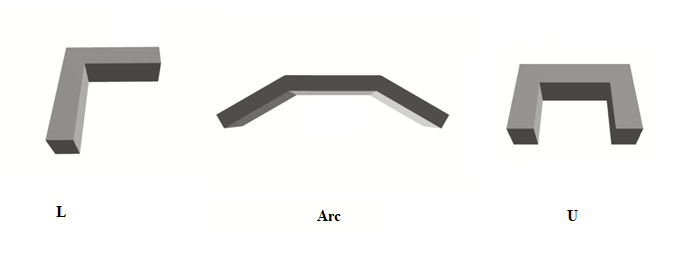}
   \caption{Experimental Objects}
   \label{fig.4}
\end{figure}

\begin{figure}[ht]
  \centering
  \includegraphics[width = 8cm]{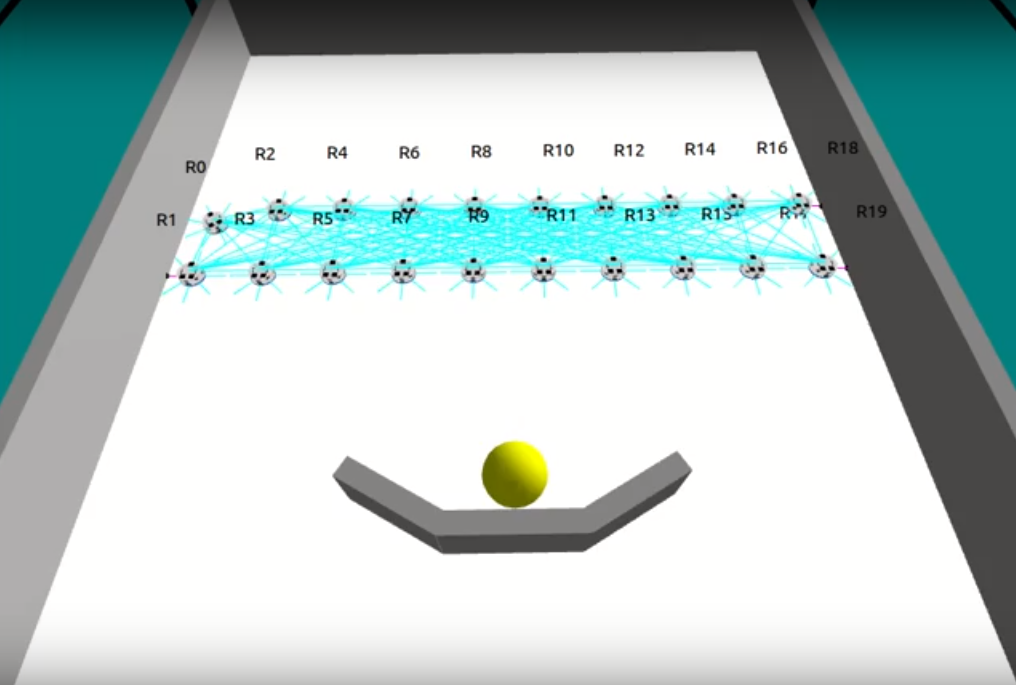}
   \caption{Experimental Setup: Proposed Filling to form Convex Object}
   \label{fig.5}
\end{figure}
\begin{figure}[ht]
  \centering
  \includegraphics[width = 8cm]{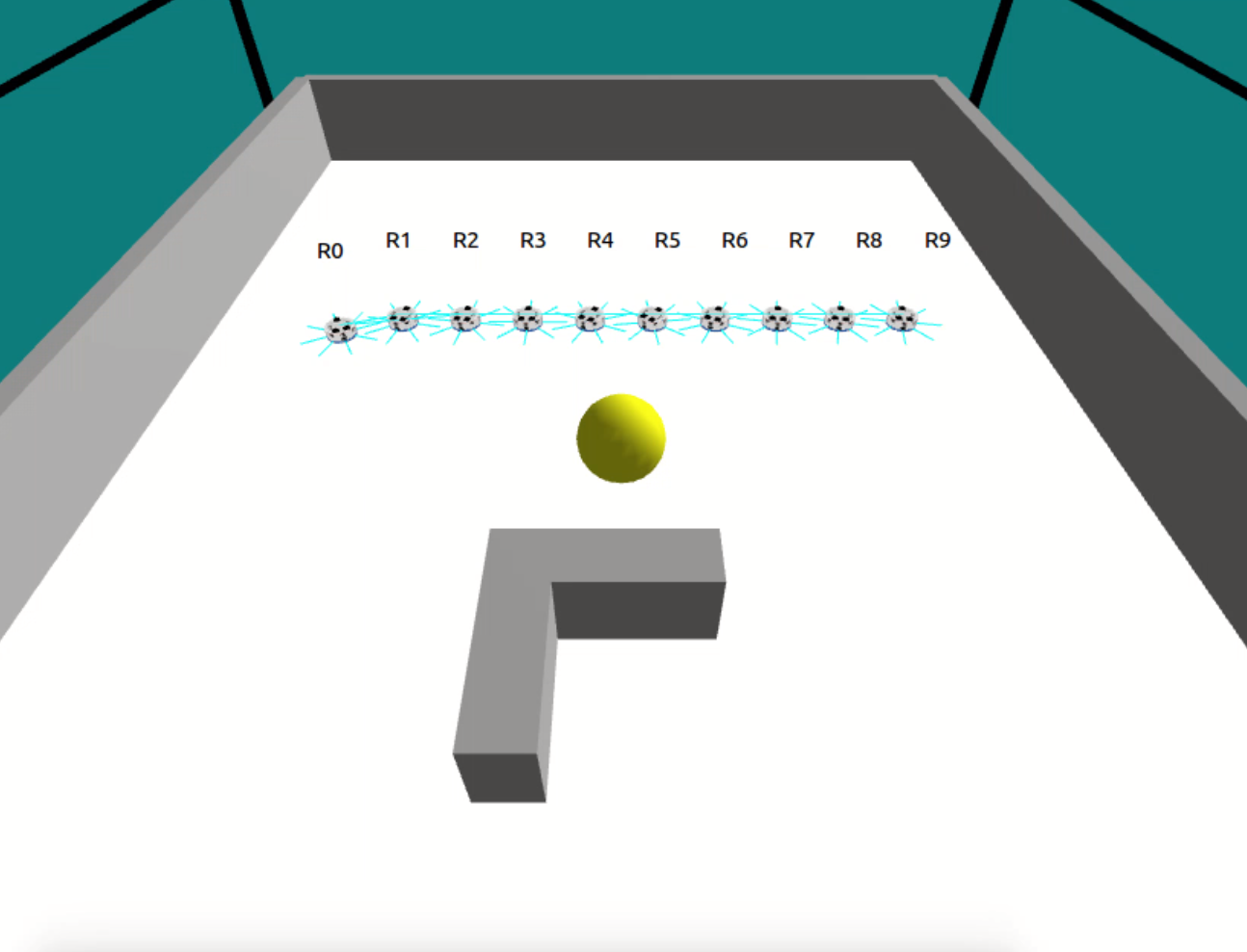}
   \caption{Experimental Setup: Proposed Filling to form Convex Object}
   \label{fig.6}
\end{figure}
\begin{figure}[ht]
  \centering
  \includegraphics[width = 8cm]{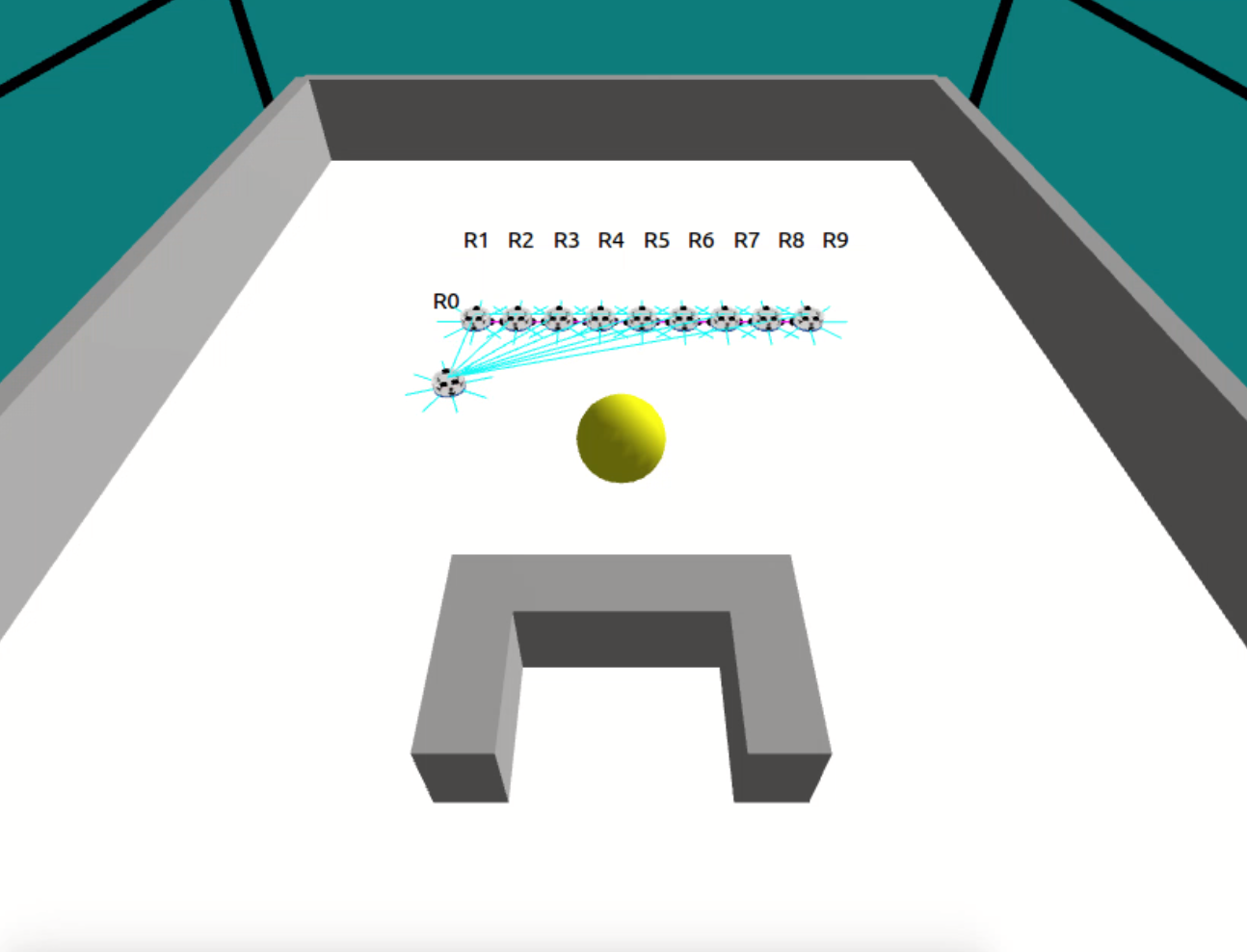}
   \caption{Experimental Setup: Proposed Filling to form Convex Object}
   \label{fig.7}
\end{figure}
We have experimented on three different shapes and size objects as shown in figure \ref{fig.4}. In order to develop concave object we have used small convex objects to form one big concave object. Using the Concave filling algorithm discussed below, we have experimented concavity filling on these different shape objects.

\subsubsection{Algorithm}
On initialization, the robots perform a random walk to look for the object and their LED's are switched off. Once in a certain range of the object, the robots turn red indicating that it has started using their proximity sensors to detect the object geometry while performing left hand wall behavior.
As soon as 4 proximity readings (theoretically it should be 5 readings) are detected simultaneously, the robot turns blue,signaling that it has turned into a 'pseudo object'.
\begin{algorithm}[h!]
\textbf{Concave Filling}\\
Rotate about own axis\\
\SetAlgoLined
  \eIf{Object not in range} {
   Perform Random walk\;
   }{
   Move to the Object\;
  }
\SetAlgoLined 
\eIf{alpha greater than pi} {
   Attach to Object\;
   Change id == Object\;
   }{
   Perform Left Hand Wall Rule\;
  }
\end{algorithm}

\subsubsection{Parameters}
We tested the following parameters for the concave filling.
\begin{itemize}
\item Distribution of the robots:\\
The distribution of the robots under a certain range slightly affected the performance of the system. It can be said that the distribution of the robots affects the concave filling but in most cases, the robots were still able to accomplish the task.
\\However, placing the robots too close to the object during initialization or orienting the robots away from the light source doesn't result in a good performance. 

\item Number of robots executing Concave filling:\\
The number of robots affect the performance in two ways.
\begin{enumerate}
\item The total number of robots executing the task determine the degree of the filling in the sense that less number of robots would be incapable of filling the object depending on it's size. Therefore, enough robots have to be deployed to completely fill the object. 

\item The number of robots simultaneously deployed affect the performance because the proximity sensors detect the neighbor robots yet do not distinguish them from the object. The robots are programmed to stop as soon as 4 sensors simultaneously detect readings.    
\end{enumerate}

\item Light Source above the object:\\
The light source above the object does not really affect the performance in a given range. The robots do not use the light source during the concave filling, they just use the light source to move towards the object. Once, in a certain range, they only use their proximity readings.
On changing the location of the light source, the robots may approach the object differently though.

\item Shape and size of the object:\\
The time and accuracy of concave filling varies according to the shape of the object. 
However, for the the same shape, it's size doesn't necessarily affect the performance drastically. The increase in size however may require additional robots.  

\end{itemize}

\subsubsection{Results}
\begin{enumerate}
\item Analysis 1: U shape as shown in figure \ref{Analysis.1}
  \begin{figure}[!ht]
    \centering
    \includegraphics[width = 8cm]{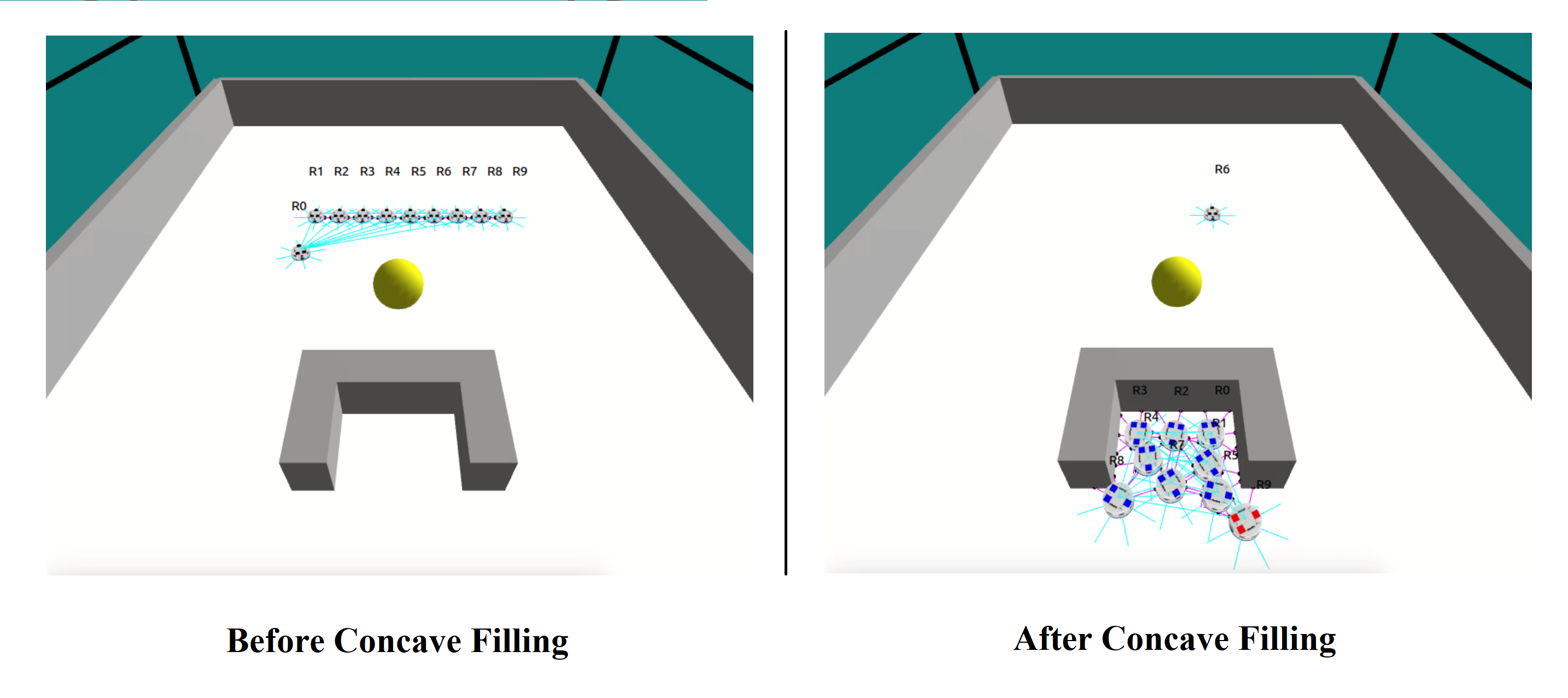}
     \caption{}
     \label{Analysis.1}
  \end{figure}
\item Analysis 2: U shape with 15 degree orientation as shown in figure  \ref{Analysis.2}
  \begin{figure}[!ht]
    \centering
    \includegraphics[width = 8cm]{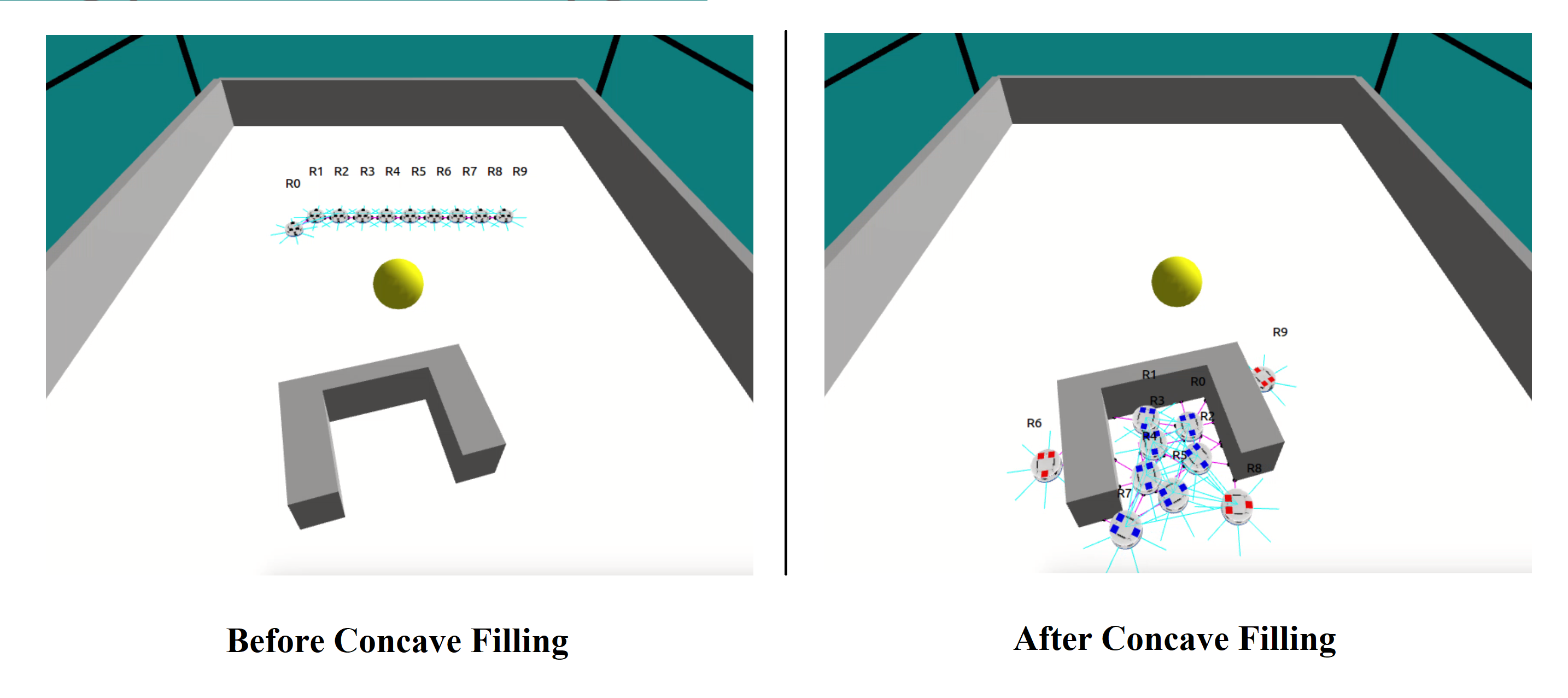}
     \caption{}
     \label{Analysis.2}
  \end{figure}
\item Analysis 3: L shape as shown in figure \ref{Analysis.3}
  \begin{figure}[!ht]
    \centering
    \includegraphics[width = 8cm]{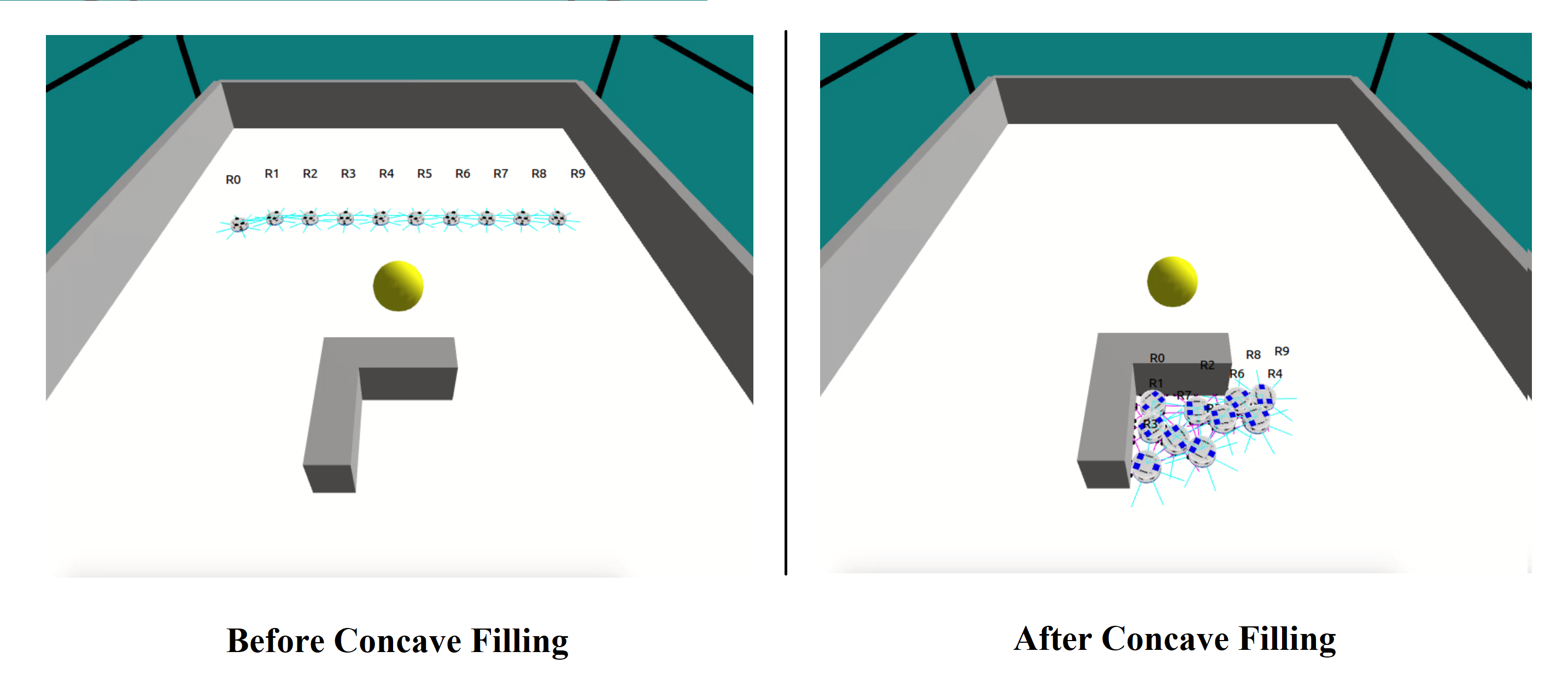}
     \caption{}
     \label{Analysis.3}
  \end{figure}
\item Analysis 4: L Shape: Changing the Robots Initial Orientation as shown in figure \ref{Analysis.4}
  \begin{figure}[!ht]
    \centering
    \includegraphics[width = 8cm]{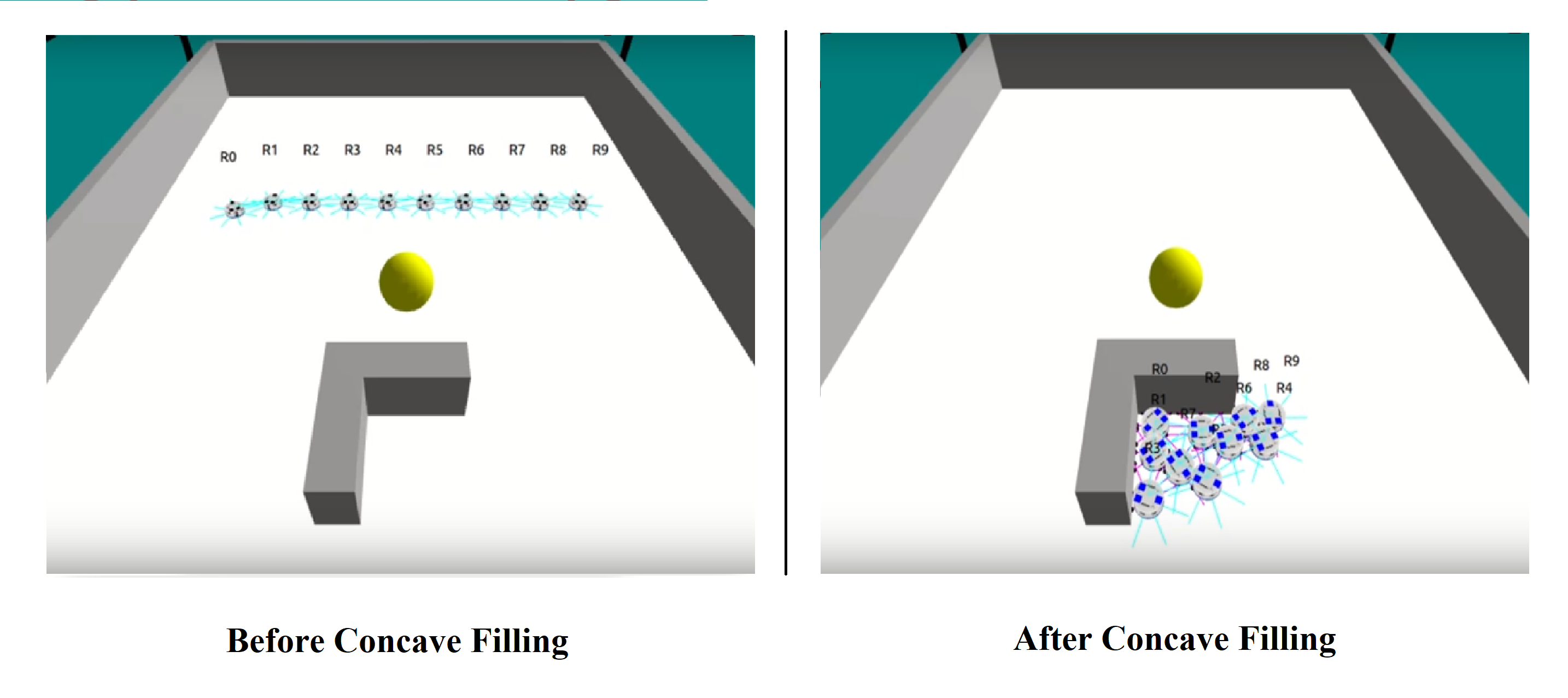}
     \caption{}
     \label{Analysis.4}
  \end{figure}
\item Analysis 5: L Shape: Changing the number of Robots as shown in figure \ref{Analysis.5}
  \begin{figure}[!ht]
    \centering
    \includegraphics[width = 8cm]{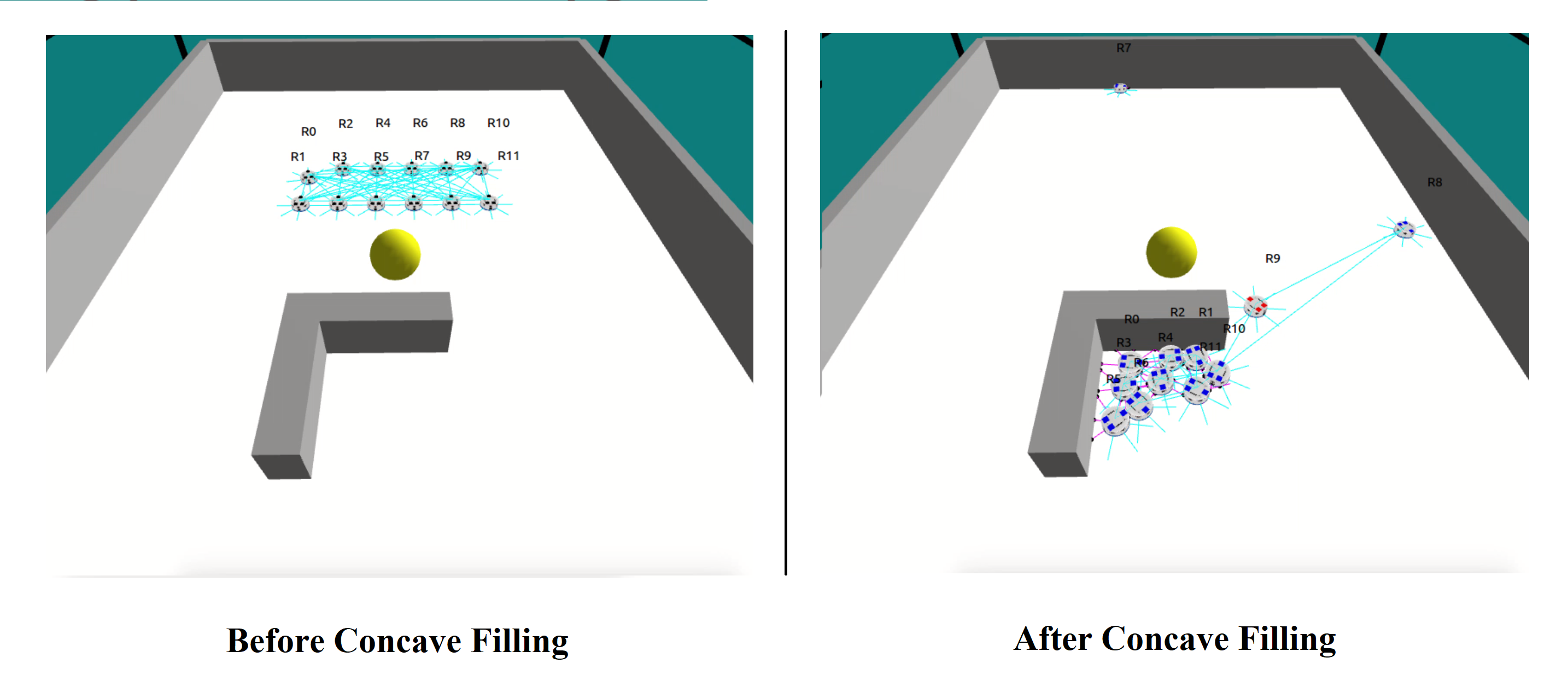}
     \caption{}
     \label{Analysis.5}
  \end{figure}
\item Analysis 6: Arc Shape as shown in figure \ref{Analysis.6}
  \begin{figure}[!ht]
    \centering
    \includegraphics[width = 8cm]{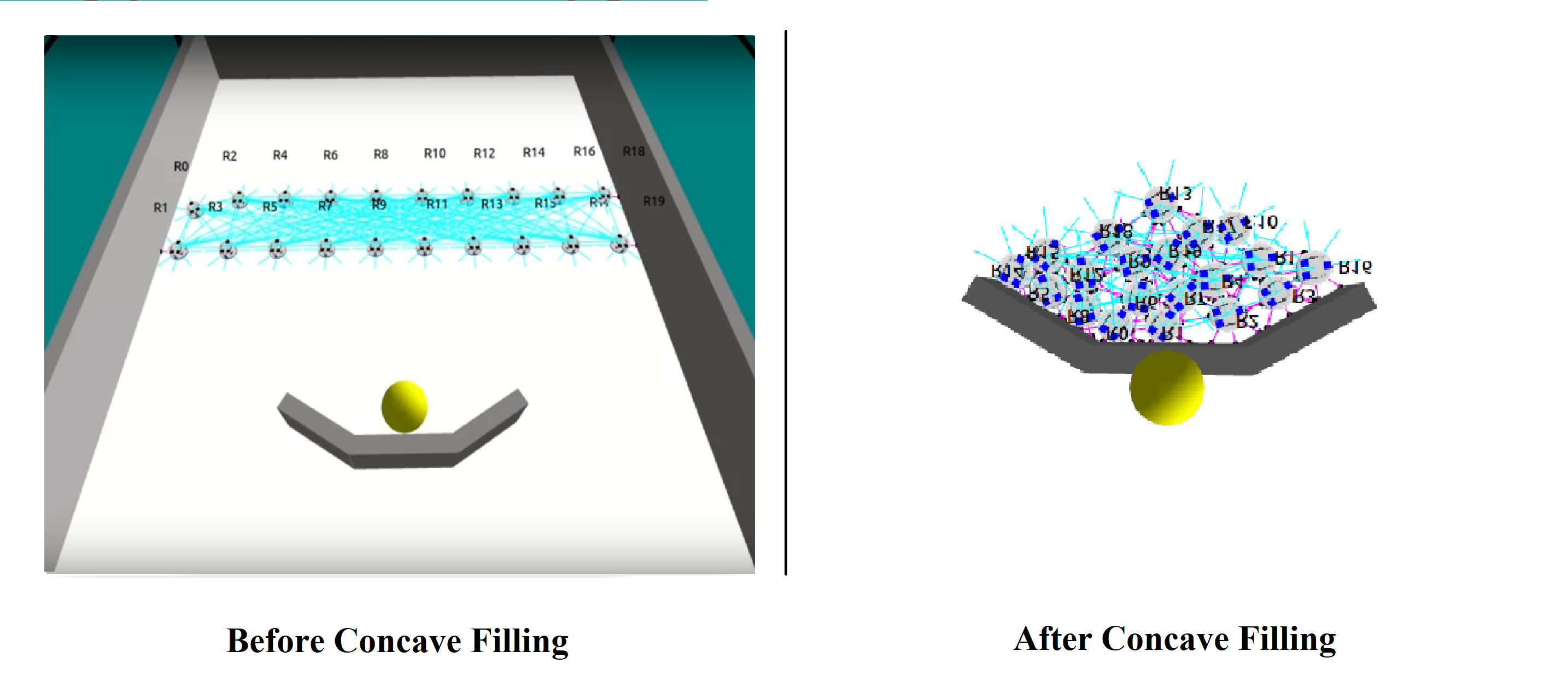}
     \caption{}
     \label{Analysis.6}
  \end{figure}
\end{enumerate}

\subsection{Experiment 2: Collective Transport}
In our second experiment we replicated the Occlusion-based convex object based experiments implemented in \cite{c1} for convex object. 

\subsubsection{Experimental Setup}
\begin{figure}[ht]
  \centering
  \includegraphics[width = 8.5cm]{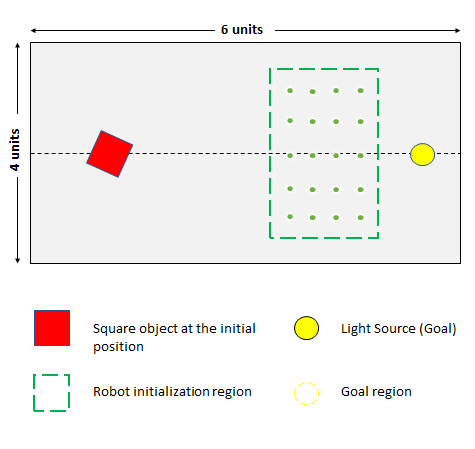}
   \caption{ Experimental setup. The initially the robots can be placed in a grid formation  or in a randomly distributed manner.}
   \label{fig.14}
\end{figure}

% \begin{figure}[h]
%   \centering
%   \includegraphics[width = 6cm]{Exp2setup.png}
%    \caption{ Experimental setup in ARGoS environment with Footbot robots: TOP VIEW}
%    \label{fig.7}
% \end{figure}

\begin{figure}[ht]
  \centering
  \includegraphics[width = 8.5cm]{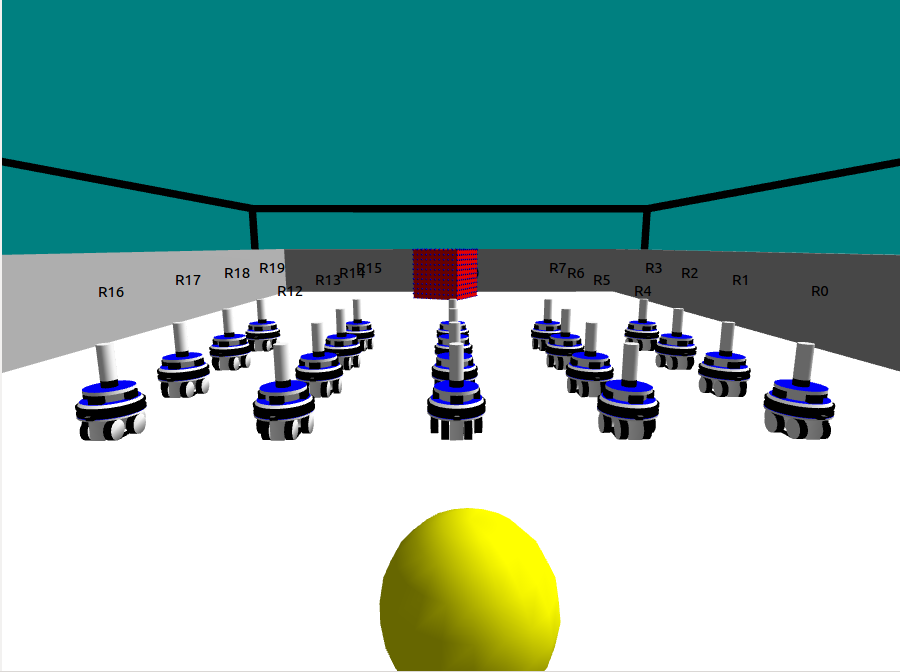}
   \caption{ Experimental setup in ARGoS environment with Footbot robots}
   \label{fig.15}
\end{figure}

The environment of the experiment as shown in figure \ref{fig.7} is a rectangular arena that is bounded by the walls. The floor of the arena has white color, and its walls are painted in gray. The goal is a light source, depicted as a yellow sphere, placed at a certain height to avoid robots collision with it.

The initial configuration of a trial is illustrated in figure \ref{fig.6}. We also have initial configurations of trials changed to randomly distributed robots in the environment. The object’s centroid and the orientation was positioned randomly as shown. The sequence of the implementation is explained with the algorithm.

\subsubsection{Algorithm}
The algorithm for Occlusion based experiment is implemented for convex objects in this paper. Further it can also be performed on concave objects after filling their concavity. The experiments starts with finding the object position. The robots in the environment also have their own 'X' and 'Y' positions along with the objects position. The object position is dynamic in nature, hence as the object moves we get its updated position. 

With the help of both these positions the robot can find the angle at which the object is placed from it. Once it gets the orientation direction of the object, it keeps checking if its current orientation is directing towards the object. If the orientation is towards the object,it updates the status of it being oriented. If not, it continues the process of rotating to orient itself. 

Once the robot is oriented, it starts moving towards the object. After reaching the object, the robots use the proximity sensors to maintain distance from the object and follow the left hand wall rule. While performing the left hand wall rule the robots keep checking for light using their light sensors. As the robot reaches the occluded region on the object it doesn't get any readings on the light sensor.

Now it is declared that the robot is in occluded area of the object. The next task is to push the object. In order to push the object, the robot again orients itself to the object following the initial orientation steps using the current position of the object and its own current position. Once oriented it starts moving towards the object, but this time it is not constrained by the proximity sensor hence it touches the object and pushes it. As the occluded region is decided by light source the robots push towards the light source which is our goal.

\begin{algorithm}[h!]
\textbf{Occlusion Based Algorithm} \\
Find Robot Position\\
Find Object Position\\

\SetAlgoLined
Calculate Orientation\\
\eIf{Orientation is true} {
   Move in Oriented Direction\;
   }{
   Change Orientation to the Object\;
  }
Reached Object \\
Perform Left Hand Wall Rule\\
\SetAlgoLined
  \eIf{ Light Source is Visible} {
   Continue Left Hand Wall Rule\;
   }{
   Rotate to align with the object \;
   Start Pushing the Object \;
  }
  
  \SetAlgoLined
  \uIf{Object touches the wall}{
    STOP \;
  }
  \uElseIf{Object reached at the Light}{
    Experiment successful - STOP \;
  }
  \Else{
    Time limit crossed - STOP \;
  }
\end{algorithm}

\subsubsection{Parameters}
For the occlusion based transport, we tested the following parameters:- 
\begin{itemize}
\item Distribution of the robots:\\
One of the most important parameters to test the robustness of the experiment is to test whether the robots are able to perform the task irrespective of how they have been placed in the environment. We carried out experiments by placing the robots randomly in the environment and then placing them in a formation as shown in figure \ref{fig.7} to compare the results obtained. When the robots were placed randomly the result were better than when placed in the formation. As robots initially try to orient themselves towards the object and then start moving towards the object the random distribution enables them to avoid other robots which have not yet being oriented hence giving better results. But when the robots are placed in formation, all the robots orient together hence start to move towards the object together. As the condition of reaching the object is analyzed based on the proximity values, the robots are misguided when they come across other robots. Hence, considering this other robot as an object and performing left hand wall rule. This happens at a large scale when the robots are placed in grid formation. 

\item Position of the object:\\
The position of the object also influences the performance efficiency of the experiment.When the object was placed near walls it was noticed that the robots approach the object but they get lost. When near the wall the robots start applying the left hand wall rule on the wall instead of the object. Hence, taking rounds along the wall of the arena. Hence, affecting the success rate of the trials. 
\end{itemize}

  \begin{figure}[!ht]
    \centering
    \includegraphics[width = 8cm]{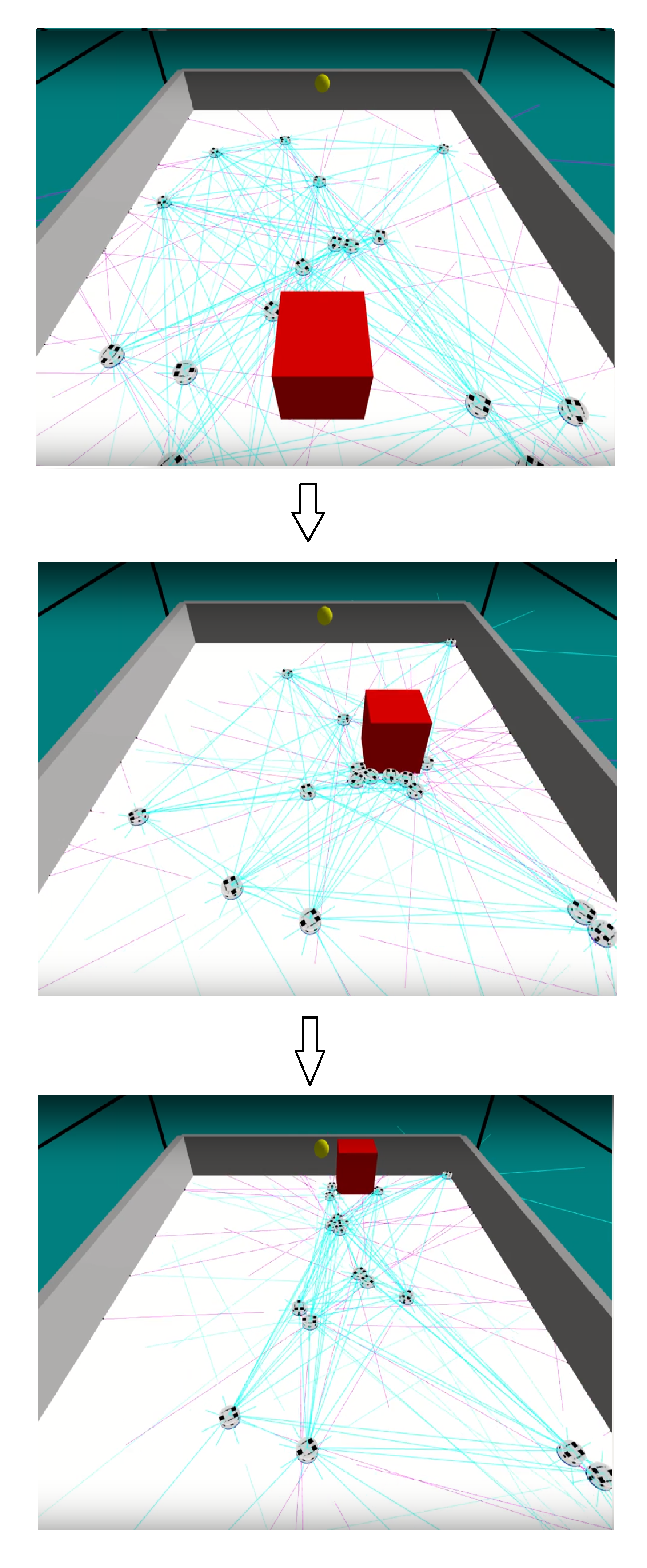}
     \caption{Occlusion Based Collective Transport of a convex object}
     \label{Analysis.7}
  \end{figure}

\section{Limitations}
\subsection{Concave Filling}
\begin{itemize}
\item The range of proximity sensors affects the performance of the experiments due to the fact that Khepera IV has a limited range. Hence, it is very hard to achieve 5 proximity readings simultaneously ( it signals that the angle is greater than \(pi\)).
\end{itemize}

\begin{itemize}
\item Simultaneously deploying too many robots affects the performance as the robots cannot distinguish between object and neighbor robots. The robots react when they come close to their neighbor as they would for the object.
\end{itemize}

\subsection{Collective Transport}
As we are not using camera on the robot we are not able to distinguish between the object to be transported and walls of the arena. Hence causing the robot to perform the left-hand wall rule along the walls of the arena. Another limitation is that objects cannot be placed near the wall. The path followed by the robots is not optimal and may vary from shape to shape.

\section{Conclusion}
The occlusion based collective transport strategy can be used to perform collective transport when the object and robot position is known. 
The concave filling algorithm presented can be used to eliminate or minimize the concavity of the object.The robots need not have any prior knowledge about the object’s geometry to perform the concave filling. The object can then be transported as a “convex” object, using the occlusion based transport strategy.

\section{Future Work}
In the future we would like to perform the same experiment using a camera. This would eliminate the use of light source being placed above the object for concave filling. Also, this would allow to differentiate robots between its neighbors, wall, object and goal.
We would also like to integrate both concave filling and occlusion based transport as a single experiment. A path optimization technique can be used so that the robots follow an optimal path. Obstacle avoidance can also be used so that the robots do not collide with its neighbors.

\bibliographystyle{IEEEtran}
\bibliography{references}

\end{document}